\documentclass[runningheads]{llncs}
\usepackage{graphicx, amsmath}
\usepackage{amssymb}
\usepackage{subcaption}
\usepackage{xcolor}
\usepackage{multirow}
\begin{document}
\title{Top-down Attention Recurrent VLAD Encoding for Action Recognition in Videos}
\titlerunning{Top-down Attention Recurrent VLAD Encoding for Action Recognition}
%
\author{Swathikiran Sudhakaran\inst{1,2} \and
Oswald Lanz\inst{1}}
\authorrunning{S. Sudhakaran and O. Lanz}
%
\institute{Fondazione Bruno Kessler, Trento, Italy \and University of Trento, Trento, Italy\\
\email{\{sudhakaran,lanz\}@fbk.eu}}

\maketitle
\begin{abstract}
	Most recent approaches for action recognition from video leverage deep architectures to encode the video clip into a fixed length representation vector that is then used for classification. For this to be successful, the network must be capable of suppressing irrelevant scene background and extract the representation from the most discriminative part of the video. Our contribution builds on the observation that spatio-temporal patterns characterizing actions in videos are highly correlated with objects and their location in the video. We propose Top-down Attention Action VLAD (TA-VLAD), a deep recurrent architecture with built-in spatial attention that performs temporally aggregated VLAD encoding for action recognition from videos. We adopt a top-down approach of attention, by using class specific activation maps obtained from a deep CNN pre-trained for image classification, to weight appearance features before encoding them into a fixed-length video descriptor using Gated Recurrent Units. Our method achieves state of the art recognition accuracy on HMDB51 and UCF101 benchmarks.
	
\end{abstract}
	\section{Introduction}
	\label{sec:intro}
	
	Despite the recent advancements in deep learning which resulted in huge performance improvements in computer vision tasks such as image recognition \cite{he2016deep}, object detection \cite{ren2017faster}, semantic segmentation \cite{Peng_2017_CVPR}, etc., video action recognition still remains a challenging task. This can be attributed mainly to two major reasons, one being the lack of large scale video datasets to enable deep networks with millions of parameters to be tuned effectively to the given task, which can be partly solved by making use of large image datasets such as imagenet for pre-training the network. The second and the most important challenge is present in the nature of the data itself, i.e., the varying duration of action instances and the huge variability of action-specific spatio-temporal patterns in videos. The former can be addressed by sampling operations such as max pooling or average pooling
	while the latter requires a careful design of network structure 
	capable of encoding the spatial information present in each frame in relation to how it evolves in subsequent frames with each action instance.
	
	Several techniques have been proposed for encoding spatio-temporal information present in videos. Simonyan and Zisserman \cite{simonyan2014two} propose to use stacked optical flow along with video frames for encoding both appearance and motion information present in the video. A huge performance improvement was observed after incorporating the optical flow stream to the image based convolutional neural network (CNN) which confirms the fact that a simple CNN has limited capability in capturing spatio-temporal information. Wang \emph{et al.} \cite{wang2015action} propose to combine improved dense trajectory method with the two stream network of \cite{simonyan2014two} to perform an effective pooling of the convolutional feature maps. The original two stream network \cite{simonyan2014two} is further improved by adding residual connections from the motion stream to the appearance stream in \cite{feichtenhofer2016spatiotemporal}. Wang \emph{et al.} \cite{wang2016temporal} propose to use several segments of the video on a two stream network and predict the action class of each segment followed by a segment consensus function for predicting the action class of the video. This enables the network to model long term temporal changes. A 3D convolutional network is proposed by Tran \emph{et al.} \cite{tran2015learning} to encode spatio-temporal information from RGB images. Carreira and Zisserman \cite{carreira2017quo} propose to combine this model with the two stream model to further improve the spatio-temporal information captured by the network. Several techniques that use recurrent neural networks such as LSTM \cite{donahue2015long,sharma2015action}, ConvLSTM \cite{sudhakaran2017convolutional,sudhakaran2017learning}, have also been proposed for encoding long term temporal changes. Girdhar \emph{et al.} \cite{girdhar2017actionvlad} propose an end-to-end trainable CNN with a learnable spatio-temporal aggregation technique.
	
	The majority of the methods mentioned above use optical flow
	images along with RGB frames for improving the recognition performance. 
	The major drawback associated with this is the huge amount of computations required for optical flow image generation. An interesting observation that we noticed is that the performance improvement brought about by the addition of the optical flow stream to these methods were almost the same ($\approx$17\% in the state of the art techniques that use a flow stream in addition to the RGB stream) \cite{simonyan2014two,wang2015action,feichtenhofer2016spatiotemporal,carreira2017quo,girdhar2017actionvlad}. This emphasizes the importance in improving the appearance stream, which uses RGB frames instead of optical flow, in order to further improve the performance of action recognition methods. 
	
\begin{figure}[t]
    \centering
    \includegraphics[width=\textwidth]{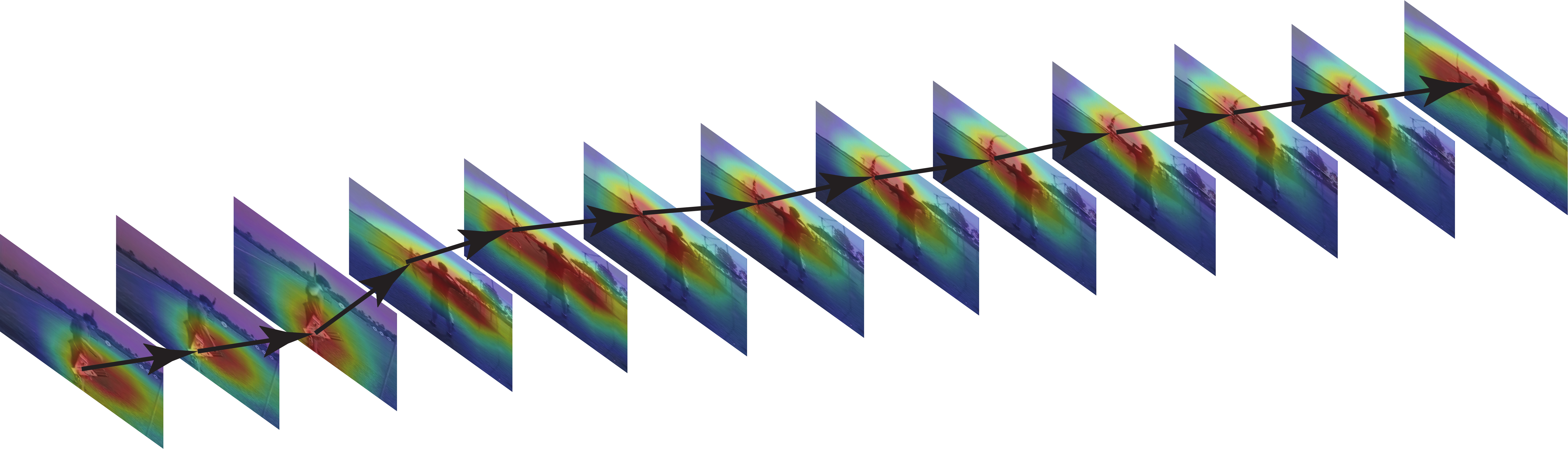}
    \caption{The figure shows the video frames taken from the action class `shoot bow'. The object that is representative of the action class, the bow, is changing its position in subsequent frames and determines the frame regions for feature extraction. In addition, the temporal order with which the image objects evolve into the action is relevant for aggregating frame features into a video descriptor.}
    \label{fig:fig1}
    \end{figure}

	In this paper, we propose to use spatial attention during the feature encoding process to address this problem. Spatial attention have been proved to be useful in several applications such as image captioning \cite{xu2015show}, object localization \cite{teh2016attention}, saliency prediction \cite{wang2017deep}, action recognition \cite{sharma2015action}. Majority of these works use bottom-up attention whereas we propose to use top-down attention. Both these attention mechanisms are used by the human brain for processing visual information \cite{ungerleider2000mechanisms}. Bottom-up attention is based on the salient features of regions in the scene such as how one region differs from another, while top-down attention uses internally guided information based on prior knowledge such as the presence of objects and how they are spatially arranged. Since majority of the actions are correlated with the objects being handled, we propose to make use of the prior information embedded in a CNN trained for image classification task for identifying the location of the objects present in the scene. The location information is decoded from the raw video frames in the form of attention maps for weighting the spatial regions that provide discriminant information in differentiating one class from another. We leverage on class activation mapping \cite{zhou2016learning} which was originally proposed for fine-grained image recognition for generating the attention map. Once the pertinent objects in the frames are identified and located, we aggregate the features from the regions into a fixed-length descriptor of the video. This is illustrated in figure \ref{fig:fig1} in which the object that is representative of the action class, the bow, is changing its position in subsequent frames. We use recurrent memory cells with parameters to learn such spatio-temporal aggregation for video classification. In summary,
	\begin{itemize}
		\item We propose Top-down Attention Action VLAD (TA-VLAD), a novel architecture that integrates top-down spatial attention with temporally aggregated VLAD encoding for action recognition from videos; 
        \item We use class specific activation maps obtained from a deep CNN trained for image classification as the spatial attention mechanism and Gated Recurrent Units for flexible temporal encoding; 
	\item We perform experimental validation of the proposed method on two most popular datasets and compare the results with state of the art.
	\end{itemize}
	
		The paper is organized as follows. TA-VLAD is presented in section \ref{sec:prop_method}, followed by our analysis of experimental results in section \ref{sec:results}. In Section \ref{sec:concl} we present our conclusions.
	
	\section{Proposed Method}
	\label{sec:prop_method}
	
    
	This section details our proposed method called Top-down Attention Action VLAD (TA-VLAD). We build our method on the recently proposed ActionVLAD \cite{girdhar2017actionvlad} for action recognition. In this, the authors develop an end-to-end trainable architecture that can perform VLAD aggregation of convolutional features extracted from a CNN. In order for the paper to be self-contained, we briefly explain VLAD encoding for action recognition and then present the details of our model. 
	
	
	\subsection{VLAD Encoding}
	\label{sub:actionVLAD}
	
	\begin{figure*}[!t]
		\centering
        \begin{subfigure}[b]{\textwidth}
		\includegraphics[width=\textwidth]{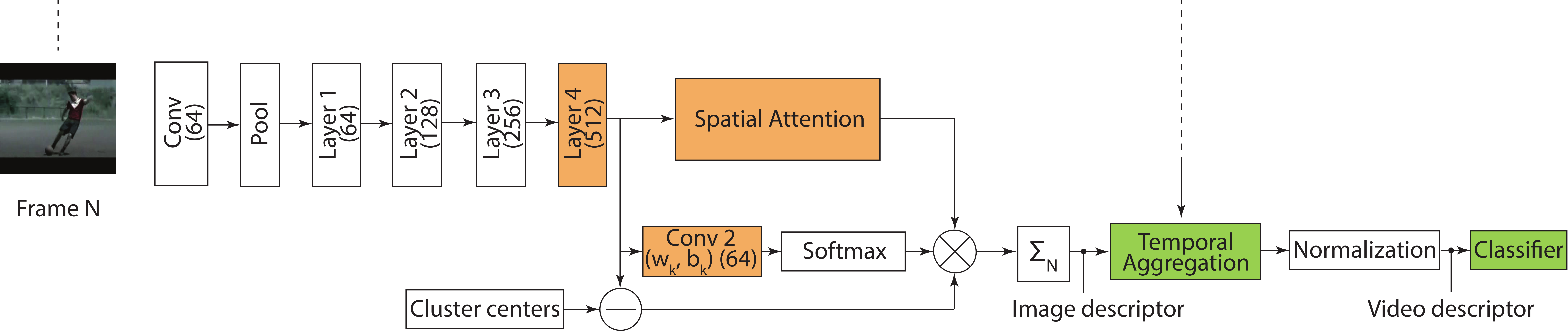}
        \caption{Proposed method}
        \label{fig:block_dia_fin}
        \end{subfigure}\\
        \vskip 5mm
        \begin{subfigure}[b]{0.6\textwidth}
\includegraphics[scale=0.16]{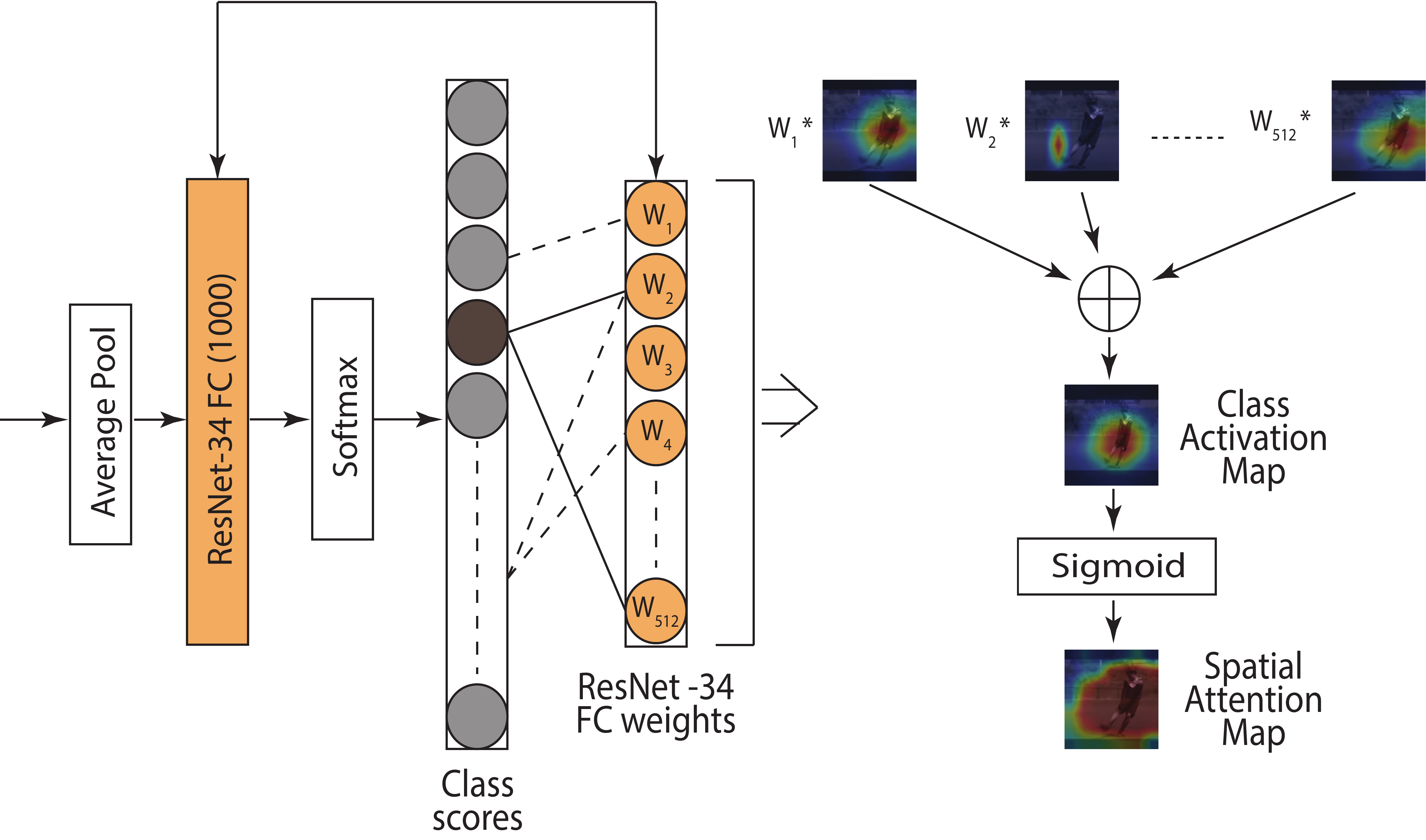}
        \caption{Spatial Attention}
        \label{fig:att_map}
        \end{subfigure}
        \hskip 4mm
        \begin{subfigure}[b]{0.34\textwidth}		\includegraphics[scale=0.14]{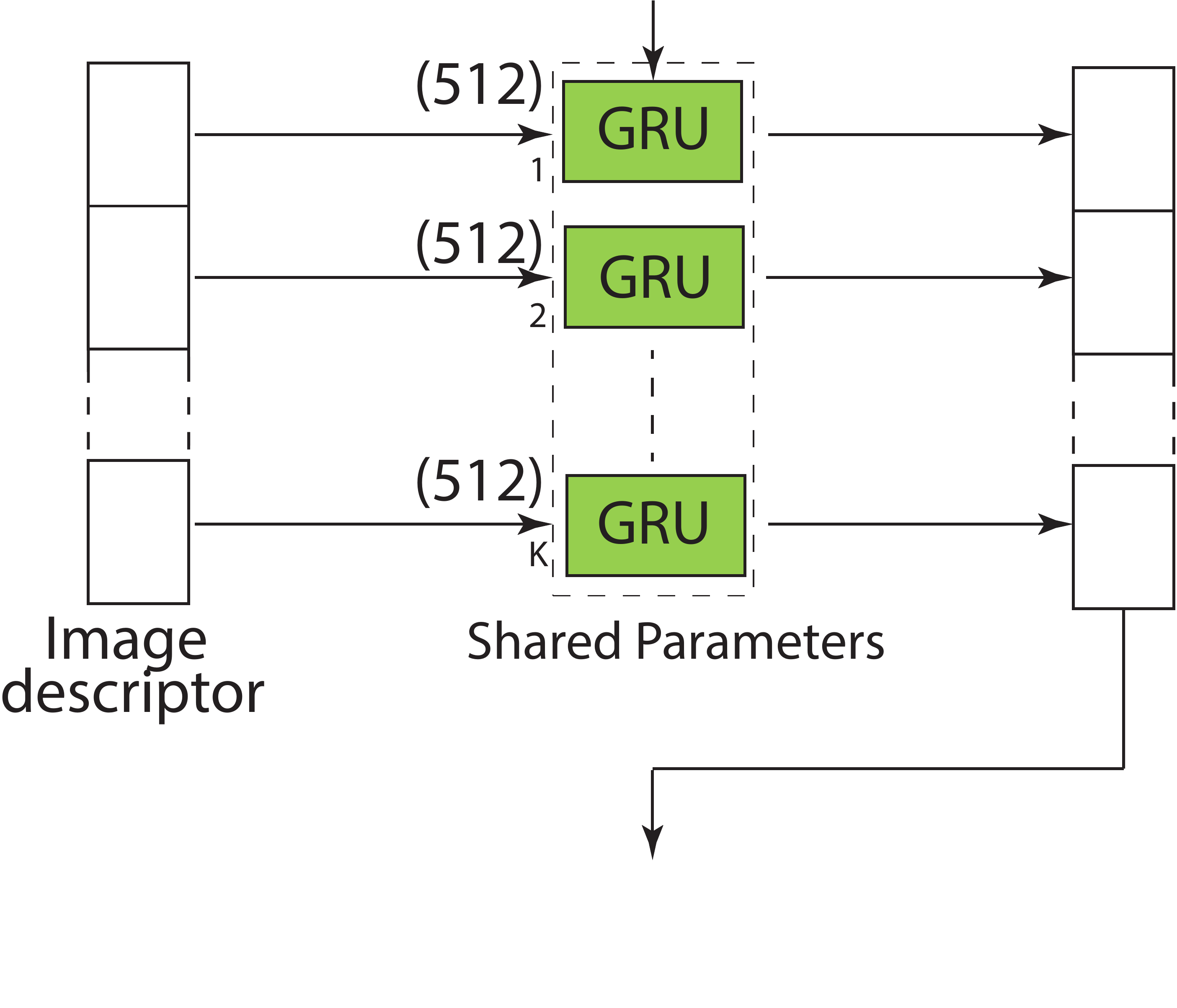}
        \caption{Temporal Aggregation}
        \label{fig:temp_agg}
        \end{subfigure}
		\caption{Block diagram of the proposed action recognition schema is given in figure (a). We use ResNet-34 for frame-based feature extraction and attention map generation, and temporally aggregated VLAD encoding to predict the action class from a frame sequence. Colored blocks indicate trained network layers, we use a two-stage strategy: stage~1 green, stage~2 green+orange. The proposed spatial attention and temporal aggregation methods are shown in figures (b) and (c) respectively.
		}
		\label{fig:block_dia}
	\end{figure*}
    
	Vector of locally aggregated descriptors (VLAD) has been originally proposed for image retrieval application \cite{jegou2010aggregating}. The method first generates a codebook of visual words $c_k$ from local feature vectors extracted from a set of training images using k-means clustering. Given a new image with local feature vectors $x_i$ of length $P$ ($i$ indexes location), the residual vectors $(x_i - c_k)$ are aggregated to form an image level descriptor $V$ using membership values $a_k(x_i)$ 
	\begin{equation}
	V(j, k) = \sum_{i=1}^{N} a_k(x_i)(x_i(j) - c_k(j))
	\label{eq:VLAD}
	\end{equation}
	where $j \in P$. The matrix $V$ is then intra-normalized, reshaped into a vector and L2-normalized to obtain the final image descriptor. A key element here is the definition of membership value $a_k(x_i)$.
	
	In the original paper \cite{jegou2010aggregating} the feature vectors were hand-crafted and aggregated with hard assignment, that is, $a_k(x_i)$ will be $1$ if $x_i$ is closest to $c_k$ and $0$ otherwise. 
	The hard assignment was later replaced with a soft-assignment in \cite{arandjelovic2016netvlad} to form an end-to-end trainable CNN-VLAD network for place recognition. Their soft-assignment membership can be conveniently described by $a_k(x_i)=softmax(W_k^Tx_i + B_k)$ where $W_k=2\alpha c_k$, $B_k=-\alpha ||c_k||^2$ with $\alpha \gg 0$ being a hyper-parameter to control selectivity (for very large $\alpha$ it approximates to hard assignment).
	
	Later Girdhar \emph{et al.} \cite{girdhar2017actionvlad} extended this method to action recognition by performing summation of the frame level descriptors, that is 
	\begin{equation}
	V(j, k) = \sum_{t=1}^T\sum_{i=1}^N a_k(x_{it})(x_{it}(j) - c_k(j))
	\label{eq:actionVLAD}
	\end{equation}
	where $T$ represents the total number of frames under consideration.
	
	Kim and Kim \cite{kim2015improving} propose an extension to the original VLAD approach by weighting the descriptors depending on their importance. For determining the spatial image locations which contain discriminant information, a saliency map of the image is used. Their method resulted in performance improvements in image retrieval application. We build upon this idea of weighted aggregation of VLAD descriptors for encoding visual features for action recognition. Instead of using a saliency detection algorithm, we develop a deep neural network with built-in weighting mechanism whose parameters are trained and which adds only minimal computation overhead to training and inference stages of the whole pipeline. Our image descriptor is computed using Eq.\eqref{eq:VLAD} with membership values
	\begin{equation}
	a_k(x_{it}) = m_{it}\cdot\,softmax(W_k^Tx_{it} + B_k)
	\label{eq:wActionVLADConv}
	\end{equation}
	where $m_{it}$ is used to weight the residual $(x_{it}-c_k)$ at each spatial location $i$. This will be detailed in sections \ref{sub:sal} and \ref{sub:wActionVLAD}.

	\subsection{Class Activation Maps for Top-Down Attention}
	\label{sub:sal}

	Saliency detection involves identifying the regions of interest present in an image. In action recognition, this constitutes the image regions where humans and the objects that they are interacting with are present. We propose to utilize the class activation map (CAM) generation technique proposed in \cite{zhou2016learning} for generating class specific saliency map of the video frames under consideration. The idea of CAMs is to project back the weights of the output layer to the convolutional feature map in the preceding layer. Let $f_l(i)$ be the activation of a unit $l$ in the final convolutional layer at spatial location $i$ and $w_l^c$ be the weight corresponding to class $c$ for unit $l$. Then the CAM for class $c$, $M_c(i)$, can be represented as
	\begin{equation}
	M_c(i) = \sum_{l}w_l^cf_l(i)
	\label{eq:CAM}
	\end{equation}
	The CAM, $M_c$, gives the regions of the image that have been used by the CNN for identifying the particular class $c$. In other words, if the classes on which the CNN has been trained is representative of the actions, then CAM gives the saliency of an image if we compute it using the winning class, i.e., the class category with the highest probability. For instance, in the case of actions such as shoot ball, draw sword, brush hair, ride bike, etc., the objects being handled by the person for performing the action gives valuable information and we propose to take advantage of this. One disadvantage of CAM is that it can be computed only on CNNs that contain a global average layer preceding the fully connected classification layer, such as ResNet \cite{he2016deep}, SqueezeNet \cite{iandola2016squeezenet}, DenseNet \cite{huang2017densely}. 
	
	Figure \ref{fig:fig_res} shows the CAM obtained for some of the frames from the videos of HMDB51 \cite{kuehne2013hmdb51} dataset (second row). We used ResNet-34 pre-trained on imagenet dataset and the class category with the highest probability is selected for the CAM computation. In the figure, the regions in the image where the objects that are part of the action are getting activated such as the ball in figures \ref{fig_ex_mod1} and \ref{fig_ex_mod2}. Thus, it can be seen that irrespective of the fact that the CNN was pre-trained on a different dataset for a different application, the CAM generated is able to identify the salient regions present in the frames, i.e., regions that provide discriminative information in identifying the action taking place in the frame.
	\subsection{Top-down Attention VLAD Encoding}
	\label{sub:wActionVLAD}
	In this section, we will detail how CAM can be used for weighting the image regions in order to improve the action recognition performance. A block diagram of the proposed approach is shown in figure \ref{fig:block_dia}. In this, a ResNet-34 CNN pre-trained on imagenet dataset is used for feature extraction $(x_{it})$ and top-down attention $(m_{it})$ computation. Feature extraction from frames and top-down attention computation can be carried out in a single forward pass across this network. The purpose of the top-down attention is to assign a higher weight to the regions present in the image that are useful for discriminating one action class from another while assigning a lower weight value to those regions that possess less discriminant information. 
	Top-down attention is obtained using the following equation
	\begin{equation}
	m_{it} = sigmoid(\sum_{l}w_l^cf_l(i))
	\label{eq:td-att}
	\end{equation}
	where $c$ is the winning class, $w_l^c$ is the weight in the final layer corresponding to class c, $f_l$ is the activation at the final covolutional layer. The $sigmoid$ function is used to map the values of $M_c(i)$ in equation \ref{eq:CAM} to the range $[0, 1]$. The sigmoid function can be represented as 
	\begin{equation}
	sigmoid(x) = \frac{1}{1+e^{-x}}
	\label{eq:sigmoid}
	\end{equation}
	The weights and bias of convolutional layer `Conv~2' is initialized with $W_k$ and $B_k$, respectively, as given in section \ref{sub:actionVLAD}. Once the attention weighted VLAD descriptor is obtained as in equation \ref{eq:wActionVLADConv}, we perform summation operation across the spatial dimension to obtain a $K\times 512$ descriptor. Girdhar \emph{et al.} performed summation across the temporal dimension (summation of image descriptors obtained from all frames) for generating the video descriptor. This simple summation operation is not capable of encoding the temporal evolution of the frame level features as it does not take it account the temporal ordering of frames. Following previous works \cite{donahue2015long,sharma2015action,sudhakaran2017convolutional}, we choose to apply an RNN for this purpose. Chung \emph{et al.} \cite{chung2014empirical} have found that both LSTM \cite{hochreiter1997long} and GRU \cite{cho2014learning} perform comparably when evaluated on the tasks of polyphonic music modeling and speech signal modeling. With GRU having the added advantage of less parameters, we decided to use GRU in order to effectively encode how the frame level features evolve as an action progresses. Our temporal aggregation method using GRU is illustrated in figure \ref{fig:temp_agg}. We use $K$ GRU modules with shared parameters (K is the number of clusters) in order to encode how the features associated with each of the clusters change with time. Once all the video frames are processed, we concatenate the final memory state of all the GRU modules to generate the temporally encoded descriptor. Then intra-normalization is applied as proposed in \cite{arandjelovic2013all} followed by reshaping operation to obtain the vector and L2-normalization to get the final feature descriptor of the input video. This is followed by a single fully-connected layer for classification.	
	
	\section{Experimental Results}
	\label{sec:results}
	
	
	The proposed method is tested on two popular action recognition datasets, namely, HMDB51 \cite{kuehne2013hmdb51} and UCF101 \cite{soomro2012ucf101} and compared against state of the art deep learning techniques. HMDB51 dataset consists of videos divided into 51 action classes collected from movies and Youtube. The dataset is composed of 6849 video clips. UCF101 dataset consists of 13320 videos collected from Youtube and has 101 action categories. For both the datasets, three train/test splits are provided by the dataset developers. Following standard practice, the recognition performance on the datasets is reported as the average of the recognition accuracy obtained on the three splits.
	\subsection{Implementation Details}
	\label{sub:imp}
	As mentioned in section \ref{sec:prop_method}, we use ResNet-34 for feature extraction and CAM computation. We use the feature map obtained from the final convolutional layer of ResNet-34 as the input features. The proposed method is implemented using Pytorch framework. The method consists of the following steps: 1) Extract convolutional features from random frames from the videos present in the training set to obtain the cluster centers. 2) Stage 1 training in which only the fully connected classifier layer and GRU network are trained while all the other parameters remain fixed. 3) Stage 2 training in which all the convolutional layers in the final block and fully-connected layer of ResNet-34, VLAD layers, GRU network and the classifier are trained. Stage 1 training acts as an initialization of the classifier and the GRU network while stage 2 training optimizes the features extracted from the frames, the clusters, and the attention to specialize to the given action recognition task. For the clustering step, we extracted features from random frames that are then clustered into 64 clusters using k-means algorithm. We chose the number of clusters based on the analysis carried out in \cite{girdhar2017actionvlad}. The network is trained for 50 epochs in stage 1 with a learning rate of $10^{-2}$ and 30 epochs in stage 2 with a learning rate of $10^{-4}$. In both stages, the learning rate is decayed by a factor of 0.5 after every 5 epochs. The network is trained using ADAM optimization algorithm with a batch size of 32. We also apply dropout at the final fully-connected layer at a rate of 0.5. The value of $\alpha$, explained in section \ref{sub:actionVLAD}, is chosen as 1000 following \cite{girdhar2017actionvlad}.
	
	We used 25 frames from each video that are uniformly sampled across time during training and evaluation. The corner cropping and scale jittering techniques proposed in \cite{wang2016temporal} is used as data augmentation during training. The center crop of frames are used for determining the action class during evaluation.	
	
		
    
    	\begin{table}[t]
		\centering
		\caption{Recognition accuracy (in \%) obtained on the first split of HMDB51 for various network configurations}
		\begin{tabular}{|c|c|}
			\hline
			\parbox{1.5in}{\centering Configuration} & \parbox{1.5in}{\centering Accuracy} \\
			\hline \hline
			ActionVLAD-Resnet34 & 53.1 \\
			\hline
			VLAD+GRU(512) & 54.4 \\
			\hline 
            VLAD+GRU(256) & 54.7 \\
            \hline
            VLAD+GRU(128) & 54.1 \\
            \hline
            TA-VLAD & 56.1\\
            \hline
		\end{tabular}
		
		\label{tab:ablation}
	\end{table}
    
		

	\subsection{Results and Discussion}
	\begin{figure*}[t]
		\centering
		\begin{subfigure}[b]{0.2\textwidth}
			\includegraphics[width=65px, height=38px]{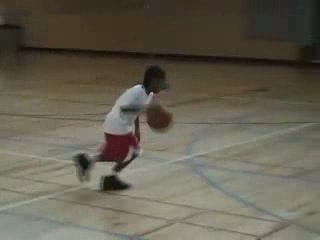}
			\label{fig_ex_img1}
		\end{subfigure}
		\begin{subfigure}[b]{0.2\textwidth}
			\includegraphics[width=65px, height=38px]{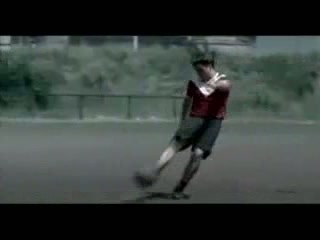}
			\label{fig_ex_img2}
		\end{subfigure}
		\begin{subfigure}[b]{0.2\textwidth}
			\includegraphics[width=65px, height=38px]{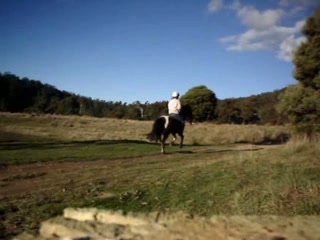}
			\label{fig_ex_img3}
		\end{subfigure}
		\begin{subfigure}[b]{0.2\textwidth}
			\includegraphics[width=65px, height=38px]{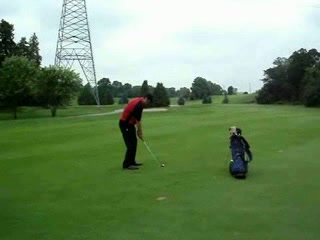}
			\label{fig_ex_img4}
		\end{subfigure}
		\\
		\vskip .5mm
		\begin{subfigure}[b]{0.2\textwidth}
			\includegraphics[width=65px, height=38px]{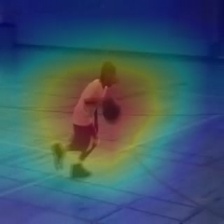}
			\label{fig_ex_org1}
		\end{subfigure}
		\begin{subfigure}[b]{0.2\textwidth}
			\includegraphics[width=65px, height=38px]{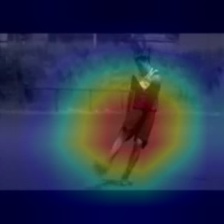}
			\label{fig_ex_org2}
		\end{subfigure}
		\begin{subfigure}[b]{0.2\textwidth}
			\includegraphics[width=65px, height=38px]{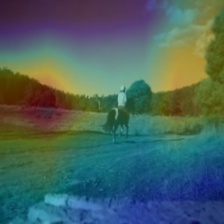}
			\label{fig_ex_org3}
		\end{subfigure}
		\begin{subfigure}[b]{0.2\textwidth}
			\includegraphics[width=65px, height=38px]{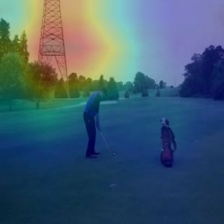}
			\label{fig_ex_org4}
		\end{subfigure}
		\\
		\vskip .5mm
		\begin{subfigure}[b]{0.2\textwidth}
			\includegraphics[width=65px, height=38px]{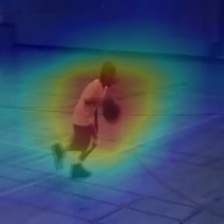}
			\caption{Dribble}
			\label{fig_ex_mod1}
		\end{subfigure}
		\begin{subfigure}[b]{0.2\textwidth}
			\includegraphics[width=65px, height=38px]{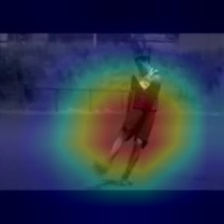}
			\caption{Kick ball}
			\label{fig_ex_mod2}
		\end{subfigure}
		\begin{subfigure}[b]{0.2\textwidth}
			\includegraphics[width=65px, height=38px]{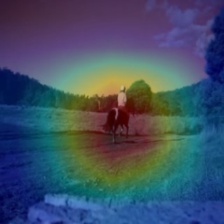}
			\caption{Ride horse}
			\label{fig_ex_mod3}
		\end{subfigure}
		\begin{subfigure}[b]{0.2\textwidth}
			\includegraphics[width=65px, height=38px]{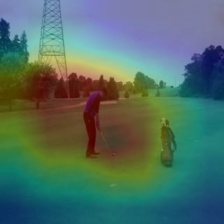}
			\caption{Golf}
			\label{fig_ex_mod4}
		\end{subfigure}\\
        \vskip 2mm
        \begin{subfigure}[b]{0.2\textwidth}
			\includegraphics[width=65px, height=38px]{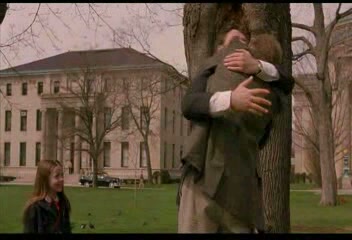}
			\label{fig_ex_img5}
		\end{subfigure}
		\begin{subfigure}[b]{0.2\textwidth}
			\includegraphics[width=65px, height=38px]{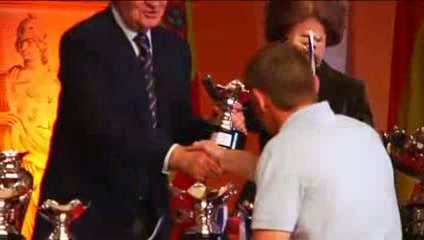}
			\label{fig_ex_img6}
		\end{subfigure}
		\begin{subfigure}[b]{0.2\textwidth}
			\includegraphics[width=65px, height=38px]{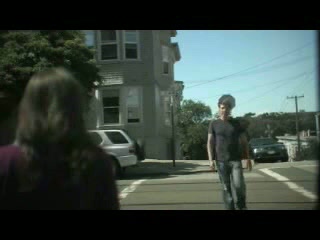}
			\label{fig_ex_img7}
		\end{subfigure}	
		\begin{subfigure}[b]{0.2\textwidth}
			\includegraphics[width=65px, height=38px]{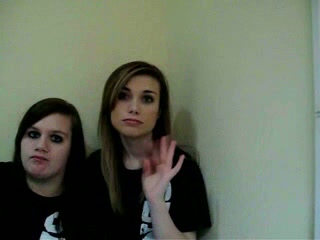}
			\label{fig_ex_img8}
		\end{subfigure}     \\
        \vskip .5mm
        \begin{subfigure}[b]{0.2\textwidth}
			\includegraphics[width=65px, height=38px]{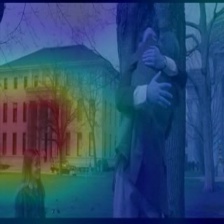}
			\label{fig_ex_org5}
		\end{subfigure}
		\begin{subfigure}[b]{0.2\textwidth}
			\includegraphics[width=65px, height=38px]{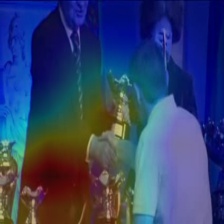}
			\label{fig_ex_org6}
		\end{subfigure}
		\begin{subfigure}[b]{0.2\textwidth}
			\includegraphics[width=65px, height=38px]{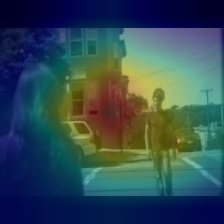}
			\label{fig_ex_org7}
		\end{subfigure}	
		\begin{subfigure}[b]{0.2\textwidth}
			\includegraphics[width=65px, height=38px]{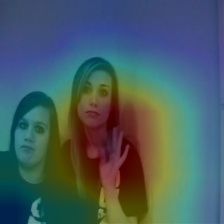}
			\label{fig_ex_org8}
		\end{subfigure}\\
        \vskip .5mm
		\begin{subfigure}[b]{0.2\textwidth}
			\includegraphics[width=65px, height=38px]{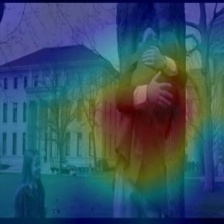}
			\caption{Hug}
			\label{fig_ex_mod5}
		\end{subfigure}
		\begin{subfigure}[b]{0.2\textwidth}
			\includegraphics[width=65px, height=38px]{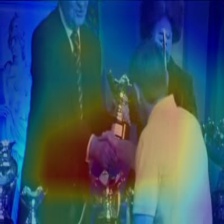}
			\caption{Shake hands}
			\label{fig_ex_mod6}
		\end{subfigure}
		\begin{subfigure}[b]{0.2\textwidth}
			\includegraphics[width=65px, height=38px]{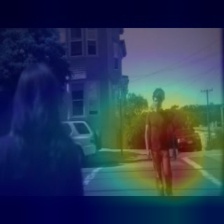}
			\caption{Walk}
			\label{fig_ex_mod7}
		\end{subfigure}	
		\begin{subfigure}[b]{0.2\textwidth}
			\includegraphics[width=65px, height=38px]{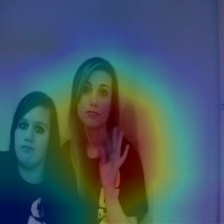}
			\caption{Wave}
			\label{fig_ex_mod8}
		\end{subfigure}
		
		\caption{Class activation maps (CAM) for some frames in HMDB51 dataset with action classes containing objects (top) and without objects (bottom). Top row: original frames, second row: CAM obtained using ResNet-34 trained on imagenet, bottom row: CAM obtained using the network trained for action recognition (after stage 2 training)}
		\label{fig:fig_res}
	\end{figure*}
	
	\begin{table*}[t]
		\centering
		\caption{Comparison of  proposed approach against state of the art methods on HMDB51 and UCF101 datasets (recognition accuracy in \%).}
		\begin{tabular}{|c|c|c|c|}
			\hline
			\parbox{1in}{\centering Method} & \parbox{1in}{\centering Backbone} & \parbox{1in}{\centering HMDB51} & \parbox{1in}{\centering UCF101} \\
			\hline  \hline
			Two-Stream VGG \cite {simonyan2014two} & VGG-M & 40.5 & 73.0\\
			\hline
			Two-Stream ResNet \cite{feichtenhofer2016spatiotemporal} & ResNet-50 &43.4 & 82.3 \\
			\hline
			TDD \cite{wang2015action} & VGG-M & 50 & 82.8\\
			\hline
			I3D \cite{carreira2017quo} & Inception V1 & 49.8 & 84.5\\
			\hline
			TSN \cite{wang2016temporal} & Inception V2 & 51 & 85.1\\
			\hline
			LSTM Soft Attention \cite{sharma2015action} & GoogleNet & 41.3 & 84.9\\
			\hline
			\hline
			ActionVLAD \cite{girdhar2017actionvlad} & VGG-16 & 49.8 & 80.3\\
			\hline \hline
			\textbf{TA-VLAD} & ResNet-34 & \textbf{54.1} & \textbf{85.7}\\
			\hline
		\end{tabular}
		\label{tab:res_table}
	\end{table*}

	We first evaluated the performance of the proposed approach on the first split of HMDB51 dataset with various settings. The results obtained are summarized in table \ref{tab:ablation}. We first adapted the ActionVLAD implementation from the authors of  \cite{girdhar2017actionvlad} using Resnet-34 for fair comparison and to establish a baseline since we are proposing to use Resnet-34 for frame level feature extraction. Then we added the proposed temporal aggregation method consisting of GRU networks. We evaluated the performance of the approach using different GRU memory size and found that a GRU network with 256 memory size gives the optimum performance. Thus we chose to use a GRU network of size 256 for temporal aggregation. By replacing the temporal summation with the proposed temporal aggregation method consisting of GRU network, we improved the performance by $1.6\%$ over ActionVLAD. This shows the importance of keeping temporal ordering during feature descriptor generation of a video. In the next step, we evaluated the performance of the proposed approach, TA-VLAD by adding the proposed top-down attention mechanism during the image level feature encoding stage and an improvement of $3\%$ is obtained over ActionVLAD.

	For validating that training the layers of the network used for CAM computation is useful, we decoupled the ResNet-34 used for generating CAM from the rest of the trainable layers. With the CAM branch not fine-tuned, a recognition accuracy of $55.3\%$ is obtained on the split 1 of HMDB51 dataset, as opposed to $56.1\%$ with the fully trained TA-VLAD. This shows that joint training, as explained in section \ref{sub:imp}, enables the network to improve the prior knowledge encoded within it, i.e., the relevance of objects and their locations in relation to the action performed. This can be interpreted from the example images shown in figure \ref{fig:fig_res}. From the figure, we can see that the network learns to attend to regions that contain discernible information about the action class. In addition, the network has adapted its attention for both actions containing objects (\ref{fig_ex_mod3} and \ref{fig_ex_mod4}) as well as actions that are performed in the absence of objects (\ref{fig_ex_mod5} and \ref{fig_ex_mod7}).
	
	Table \ref{tab:res_table} compares the proposed approach, TA-VLAD, with state of the art techniques. The results are the average of the recognition accuracy in $\%$ obtained over all three splits. For fair comparison, we report the performance of the compared methods using RGB frames only and consider those methods that are based on deep learning and use imagenet for pre-training. For each of the three splits in HMDB51, we obtained recognition accuracies of $56.1\%$, $52.4\%$ and $53.7\%$ and on UCF101 dataset, $85.9\%$, $85.7\%$ and $85.4\%$. From the table, it can be seen that the proposed approach performs better than the state of the art deep learning methods for action recognition.
	
	\section{Conclusions}
	\label{sec:concl}
	We presented a novel end-to-end trainable deep neural network architecture for video action recognition which makes use of top-down attention mechanism for weighting spatial regions that possess discriminant information regarding the action class. For this, we used class activation maps generated from a network pre-trained on imagenet. Experiments show that prior information about the objects present in the scene, which is applied as top-down attention, improves recognition performance of the network. We also developed a temporal aggregation scheme that encodes frame level features into a fixed length video descriptor using a GRU network that inherits the cluster structure of the feature space. The boost in the performance obtained shows the importance in considering the temporal ordering of video frames during the feature encoding process. The proposed method is tested on two most popular action recognition datasets and achieves state of the art performance in terms of recognition accuracy. In addition, it was also found that the network is able to improve the prior knowledge about the scene when the top-down attention generation network is trained jointly with the video descriptor generation network. In order to improve the performance, existing approaches use optical flow as a second modality to explicitly encode motion changes and as a future work, we will explore the possibility of adding attention mechanism to the flow modality to further improve the performance of our method.
	
\bibliographystyle{IEEEbib}
\bibliography{refs}

\end{document}